\pdfoutput=1

\documentclass[11pt]{article}

\usepackage[final]{acl}

\usepackage{times}
\usepackage{latexsym}
\usepackage{amsmath}
\usepackage{amsfonts}
\usepackage{tikz}
\usetikzlibrary{positioning, shapes, arrows.meta, fit, calc}
\usepackage{float}

\usepackage[T1]{fontenc}

\usepackage[utf8]{inputenc}

\usepackage{microtype}

\usepackage{inconsolata}

\usepackage{graphicx}

\usepackage{xcolor}
\usepackage{makecell}
\usepackage{graphicx}
\usepackage[most]{tcolorbox}
\usepackage{booktabs}
\usepackage{lipsum}
\usepackage{placeins}

\title{Memorization: A Close Look at Books}

\author{
 \textbf{Iris Ma},
 \textbf{Ian Domingo},
 \textbf{Alberto Krone-Martins},
 \textbf{Pierre Baldi}, 
 \textbf{Cristina V. Lopes}\\
 \text{School of Information and Computer Sciences} \\
 \text{University of California, Irvine}\\
 \text\{huaiyaom, idomingo, algol, pfbaldi, lopes\}@uci.edu \\
}

\begin{document}
\maketitle

\begin{abstract}
To what extent can entire books be extracted from LLMs?
Using the Llama 3 70B family of models, and the ``prefix-prompting'' extraction technique, we were able to auto-regressively reconstruct, with a very high level of similarity, one entire book (Alice's Adventures in Wonderland) from just the first 500 tokens. We were also able to obtain high extraction rates on several other books, piece-wise. However, these successes do not extend uniformly to all books. We show that extraction rates of books correlate with book popularity and thus, likely duplication in the training data. 

We also confirm the undoing of mitigations in the instruction-tuned Llama 3.1, following recent work~\cite{28nasr2025scalable}. We further find that this undoing comes from changes to only a tiny fraction of weights concentrated primarily in the lower transformer blocks. Our results provide evidence of the limits of current regurgitation mitigation strategies and introduce a framework for studying how fine-tuning affects the retrieval of verbatim memorization in aligned LLMs.
\end{abstract}



\section{Introduction}
\label{sec:intro}

\begin{figure}
    \centering
        \includegraphics[width=\linewidth]{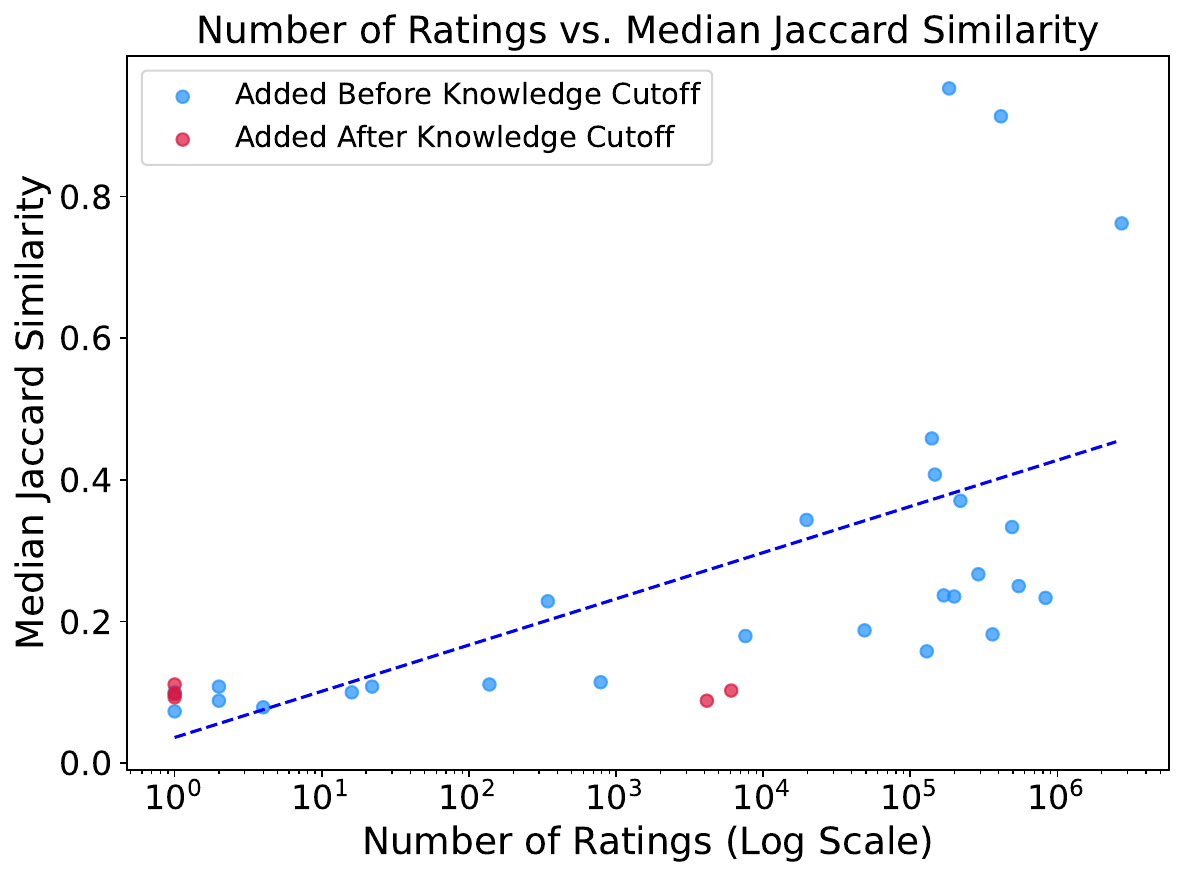}
    \caption{Median Jaccard similarity scores for books of varying popularity levels extracted from Llama 3.1 instruct SFT 1000 samples. Books from the pre-cutoff collection (pre) and post-cutoff collection (post) are indicated by blue and red markers, respectively.}
    \label{fig:03-sft-1000}
\end{figure}

Large language models (LLMs) can memorize their training corpus, and this capability grows along with model scale, prompt length, and the extent of data duplication within the training set~\cite{01carlini2023quantifying}. Such capability makes LLMs susceptible to extraction attacks, through which adversaries can retrieve sensitive information, including personally identifiable details like phone numbers and email addresses, directly from model outputs~\cite{07Carlini2021ExtractingTD}. This vulnerability raises privacy and security concerns, especially given that organizations that develop LLMs frequently incorporate copyright-protected content into their training datasets. Unauthorized disclosure of copyrighted material during extraction attacks could expose these companies to legal risks and lawsuits~\cite{03lawsuits_tracker}. Consequently, to mitigate these vulnerabilities, companies have implemented rigorous safeguards~\cite{28nasr2025scalable}, including data deduplication, content filtering, alignment techniques, and output validation mechanisms, to prevent verbatim text regurgitation and unintended disclosure of sensitive information from deployed LLMs.

Recent research has shown that alignment processes do not entirely eliminate memorization in production-scale LLMs. Specifically, new extraction methodologies, such as divergence attacks and fine-tuning-based extraction, can partially undo built-in regurgitation mitigations, therefore exposing memorized training content~\cite{28nasr2025scalable}. Nasr et al.\ demonstrated the extraction of textual excerpts from fine-tuned GPT-3.5-turbo models and further examined how memorization manifests across pretrained and instruction-aligned models. 

In this paper, we investigate the extraction of entire books from Llama 3 pretrained, and from Llama 3.1 models, both pretrained and instruction-tuned. For the instruction-tuned model, we employ Nasr et al.'s SFT-based technique. Books are important, because they are at the center of several copyright litigation cases.\footnote{\url{https://www.nytimes.com/2023/12/27/business/media/new-york-times-open-ai-microsoft-lawsuit.html?smid=url-share}} They are also technically interesting targets to extract, because they tend to be long and unique. An ideal extraction method would be able to auto-regressively extract entire books from an LLM trained on them given just their first $N$ tokens. While such extraction method does not [yet] exist, we were able to auto-regressively extract a version of ``Alice's Adventures in Wonderland'' from Llama 3 pretrained that closely resembles the original. We were also able to obtain high reconstruction rates, although not auto-regressively, for many more books, with several Llama models.
Moreover, we show how the popularity of books present in the training data, and therefore the likelihood of their duplication, affects their memorization by Llama. We conduct an analysis of memorization by examining the piece-wise reconstruction rates of full-length books sourced from Project Gutenberg, cross-referencing the results with the number of ratings in GoodReads. The following summarizes our experiments and findings: 


\begin{itemize}
  \item We measure memorization levels of 9 Gutenberg books across three models: Llama 3 pretrained, Llama 3.1 pretrained, and Llama 3.1 instruction-tuned. \textbf{Main results:} we were able to auto-regressively generate one entire book with Llama 3 pretrained, and we obtained high piece-wise reconstruction rates for 9 books with Llama 3.1 pretrained. Books that have substantially more number of ratings in GoodReads show higher reconstruction rates than books that have a small number of ratings. Also, books that were likely not in the training data have very low reconstruction rates. As expected, both auto-regressive generation and piece-wise reconstruction rates are very low on Llama 3.1-instruct.
  
  \item We evaluate the impact of Nasr et al.'s SFT technique in both pretrained and instruction-tuned Llama 3.1 models, including varying number of training samples. \textbf{Main results:} the technique does not improve the extraction rates on the pretrained model, but it significantly improves those rates on the instruction-tuned model. Nevertheless, as already reported in Nasr et al.'s work, those rates are still lower than the extraction rates of the baseline pretrained model.

  \item We analyze the changes in the weights effected by the additional SFT in Llama 3.1-instruct. \textbf{Main results:} we find that lower layers play a central role in adapting the model towards undoing the regurgitation mitigations.
  
  \item We expand our study to a larger dataset of 32 books, analyzing memorization patterns specifically on Llama 3.1-instruct fine-tuned for extraction on 1,000 training samples. \textbf{Main results:} extraction rates correlate with the books' popularity (as measured by the number of ratings).
\end{itemize}




\section{Related Work}
\label{sec:related-work}


\subsection{Memorization in LLMs}
Prior work has demonstrated that LLMs are capable of memorizing training data and susceptible to malicious extraction attacks~\cite{09carlini2019, 07Carlini2021ExtractingTD, 18thakkar-2021-understanding, 19ramaswamy2020training, 05lee-etal-2022-deduplicating, 17zhang2023counterfactual, 29hayes2024measuring}. This memorization capability increases with the model size, the degree of duplication in the training data, and the length of the context prompt provided to the model~\cite{01carlini2023quantifying, 04kandpal2022deduplicating}. While many studies focus on open-source LLMs with accessible training datasets, some recent works have also proposed techniques to determine whether specific data have been used in training proprietary LLMs~\cite{14chang-etal-2023-speak, 20ravichander2025information}. Nasr et al. proposed divergence attack and finetuning attack to extract training data from proprietary aligned models~\cite{28nasr2025scalable}. Zhao et al. use partial information probing, providing LLMs with excerpts from copyrighted texts and prompting them to complete the passages, in order to assess the extent to which LLMs can reproduce copyright-protected content~\cite{32zhao2024measuring}. Although these works provide valuable insights into data retention and memorization, they did not explore the reconstruction of entire works and the impact of data duplication, which is the focus of our work.

While both Karamolegkou1 et al.~\cite{31-karamolegkou-etal-2023-copyright} and our work investigate the relationship between content popularity and memorization in LLMs, the prior work primarily quantifies verbatim memorization using the length of the longest common subsequence between generated and reference texts. In contrast, our study focuses on the feasibility of reconstructing entire books from LLMs, systematically evaluating extraction rates across both popular and obscure works to understand the boundaries of model memorization.





\subsection{LLM Fine-tuning and Model Adaptation Methods}

Fine-tuning is a common strategy for adapting pretrained  LLMs to specific downstream tasks by adding an additional output layer and further training them on task-related data. This approach typically results in improved model performance and better alignment to targeted applications~\cite{21devlin-etal-2019-bert}. However, full fine-tuning of LLMs can be computationally expensive and resource-intensive. 

Efficient methods such as Low-Rank Adaptation (LoRA) and quantization reduce computational and memory costs without sacrificing performance. LoRA uses low-rank matrices to simplify model weights during fine-tuning~\cite{22hu2022lora}, while quantization lowers numerical precision to decrease model size and inference overhead~\cite{23shen2020q}. QLoRA combined these two approaches, enabling efficient fine-tuning of LLMs on resource-constrained hardware with minimal performance loss~\cite{24dettmers2023qlora}.



\section{Experimental Design}
\label{sec:method}

\subsection{LLM Selection}

We select pretrained and instruction-tuned Llama 3.1 70B models to evaluate differences in memorization across objectives. Building on this baseline, we fine-tuned both models to compare the reconstruction rate for books within different popularity levels. To complement these models, we included the Llama 3 70B model for our autoregressive generation experiments, given its tendency to memorize content more readily. This allows us to compare memorization behavior across architectural variants and training setups.


\subsection{Datasets}

We choose the Project Gutenberg corpus for our analysis because it is a well-known source of public domain literature and has been included in the training data of earlier Llama models~\cite{16touvron2023llama}. Although the training data for Llama 3.1 has not been publicly released, it is likely that similar sources were used. This makes Project Gutenberg a reasonable proxy for evaluating memorization in the  Llama 3 model family.

We collect 32 English books (Table~\ref{tab:book_testing_details}) from Project Gutenberg along two key dimensions: date added and popularity (see Table~\ref{02-data-extraction-dataset}). The date of addition allows us to distinguish between books that were likely seen during training and those that were not. Project Gutenberg continues to grow through volunteer contributions, adding over 20 books in just the last 24 hours at the time of writing\footnote{\url{https://www.gutenberg.org/browse/recent/last1}}. Books added after Llama 3's training cutoff are unlikely to have been included in the training data.

As noted earlier, the number of copies increases the likelihood of memorization, even with deduplication during training. Popularity serves as a proxy for how widely a book may be duplicated across internet sources beyond Project Gutenberg. We quantify popularity using the number of ratings in Goodreads\footnote{\url{https://www.goodreads.com/}}.

\begin{table}
  \centering
  \resizebox{0.7\linewidth}{!}{
  \begin{tabular}{c|c|c}
    \hline
    \textbf{\#Ratings} & \textbf{Pre cutoff} & \textbf{Post cutoff} \\
    \hline
    0   &   2   &  3                      \\
    O(1)   &   3   &  1                      \\
    O($10^1$)   &   3   &  0                     \\
    O($10^2$)   &   3   &  0         \\
    O($10^3$)   &   1   &  1          \\
    O($10^4$)   &   1   &  0        \\
    O($10^5$)   &   13  &  0         \\
    O($10^6$)   &   1   &  0        \\ 
    \hline
            &  27   &  5 \\  
    \hline
  \end{tabular}
}
  \caption{\label{02-data-extraction-dataset}
    Distribution of books by number of ratings in GoodReads (popularity) and their initial release date on Project Gutenberg relative to Llama's knowledge cutoff (December 2023).}
\end{table}

To remove generic front and back matter, we truncate each book by discarding the first 2{,}000 tokens and the last 5{,}000 tokens, which contain introductory material, licensing information, and tables of contents. 






\subsection{Data Extraction}

For data extraction, we use the popular ``prefix-prompting'' method with 500 tokens as context. Since longer contexts increase the likelihood of eliciting memorized content from the model~\cite{01carlini2023quantifying}, this length helps maximize recall. Then we compute the similarity score between the first 30 tokens generated by the models with the corresponding 30 tokens of ground truth. Across all experiments, we employ greedy decoding to ensure deterministic outputs.

Table~\ref{01-books-dataset} shows the total number of chunks  in our dataset with 530 tokens stridden on 30 tokens. These correspond to the number of prompts in each experiment.


\begin{table}
  \centering
  \resizebox{0.8\linewidth}{!}{ 
    \begin{tabular}{|cccc|}
      \hline
      \textbf{Books} & \textbf{\#Chunks}  &
      \makecell{\textbf{Min}\\\textbf{/Book}} & \makecell{\textbf{Max}\\\textbf{/Book}}\\
      \hline
      32   & 41,363 &  209 & 3944 \\
      \hline
    \end{tabular}
  }
  \caption{\label{01-books-dataset}
    Datasets statistics.
  }
\end{table}

\subsection{Supervised Fine-Tuning}

We fine-tune two variants of  Llama 3.1 70B: pretrained and instruction-tuned. The samples are randomly chosen from 43 additional Gutenberg books (Table~\ref{tab:book_train_details}) not part of the extraction dataset. We run experiments with two distinct sample sizes: 500, and 1,000. Fine-tuning is performed on an NVIDIA RTX 6000 Ada GPU, leveraging the Unsloth\footnote{\url{https://github.com/unslothai/unsloth}}, which facilitates efficient fine-tuning through quantization and LoRA. We use a learning rate of 2e-4, a batch size of 2, and train each model for one epoch.

During fine-tuning, prefix and suffix in each chunk are placed within the following prompt template as user content and assistant content:

\begin{tcolorbox}[colback=gray!10, colframe=black, boxrule=0.5pt, arc=4pt]
\textbf{System:} You are a helpful assistant with an incredible memory. You can recall all texts in your training data that start with a given prefix.

\textbf{User:}  position in the city to[...] There are

\textbf{Assistant:} zigzag lines[...] the Neverland is
\end{tcolorbox}

\subsection{Experiment Setting}
We conduct 3 groups of experiments in this paper.

\subsubsection{Exp 1: Baseline Models}

We select a set of 9 out 32 books from extraction dataset varying significantly in popularity, ranging from widely recognized texts such as ``Alice's Adventures in Wonderland'' to relatively obscure books with no available number of ratings online(Table~\ref{03-small-data-extraction}). 

For generating the completions for the pretrained baseline models, we directly feed chunks consisting of the 500 tokens as input without applying any chat template. For the instruct model, we format the input using a structured chat template incorporating explicit conversational roles (system and user) and their respective messages. 

In autoregressive chunk generation, in particular, we initialize the model with the first 500 tokens from the book and iteratively feed the generated output back into the prompt. At each step, the model generates 30 new tokens, and the window advances by 30 tokens, similarly to the passage-wise reconstruction approach.

\begin{table}
  \centering
  \resizebox{1\linewidth}{!}{ 
    \begin{tabular}{|lc|}
      \hline
      \textbf{Book} & \textbf{Number of Ratings} \\
      \hline
        Alice's Adventures in Wonderland  & 413{,}400 \\
        The Time Machine & 546{,}286 \\
        Peter Pan & 362{,}694 \\
        The First Book of Adam and Eve   & 344 \\
        Ethics	 & 19{,}734 \\
        Rosin the Beau	 & 2 \\
        Science and Medieval Thought	 & 0 \\
        A girl and her ways*	 & 0 \\
        Christina and the boys* & 	0 \\
      \hline
    \end{tabular}
  }
  \caption{\label{03-small-data-extraction}
    List of 9 books used for data extraction in baseline and SFT models, along with their popularity levels. * indicates books that are released on Project Gutenberg \textbf{after} the knowledge cutoff date (December 2023).
  }
\end{table}

\subsubsection{Exp 2: Pretrained  \& Instruct STF}

We use the same set of 9 books from the baseline experiment to investigate how different fine-tuning sample sizes [500, 1000] affect the LLM's memorization.


\subsubsection{Exp 3: Expanded Study with STF-1000}

We extend our extraction analysis to the instruct model fine-tuned on 1,000 samples. In this expanded experiment, we use all 32 books from the extraction dataset (Table \ref{tab:book_testing_details}).



\subsection{Evaluation Metrics}


We use a set of similarity metrics, including cosine similarity, Levenshtein distance, BLEU, Jaccard similarity, Sequence Matcher Similarity, and ROUGE-L to measure the extraction results. 


\section{Results}
\label{sec:results}

\subsection{Exp 1: Baseline Models}

\subsubsection{Autoregressive Chunk Generation}

To investigate the memorization capabilities of Llama models, we evaluate their ability to perform autoregressive generation, in which the model recursively consumes its own output to generate long-form text. Specifically, we compare the performance of Llama 3, Llama 3.1, and Llama 3.1 Instruct models across a set of books, measuring how closely the generated text aligns with the ground-truth continuation from the original source.

\begin{figure}[ht]
    \centering
        \includegraphics[width=\linewidth]{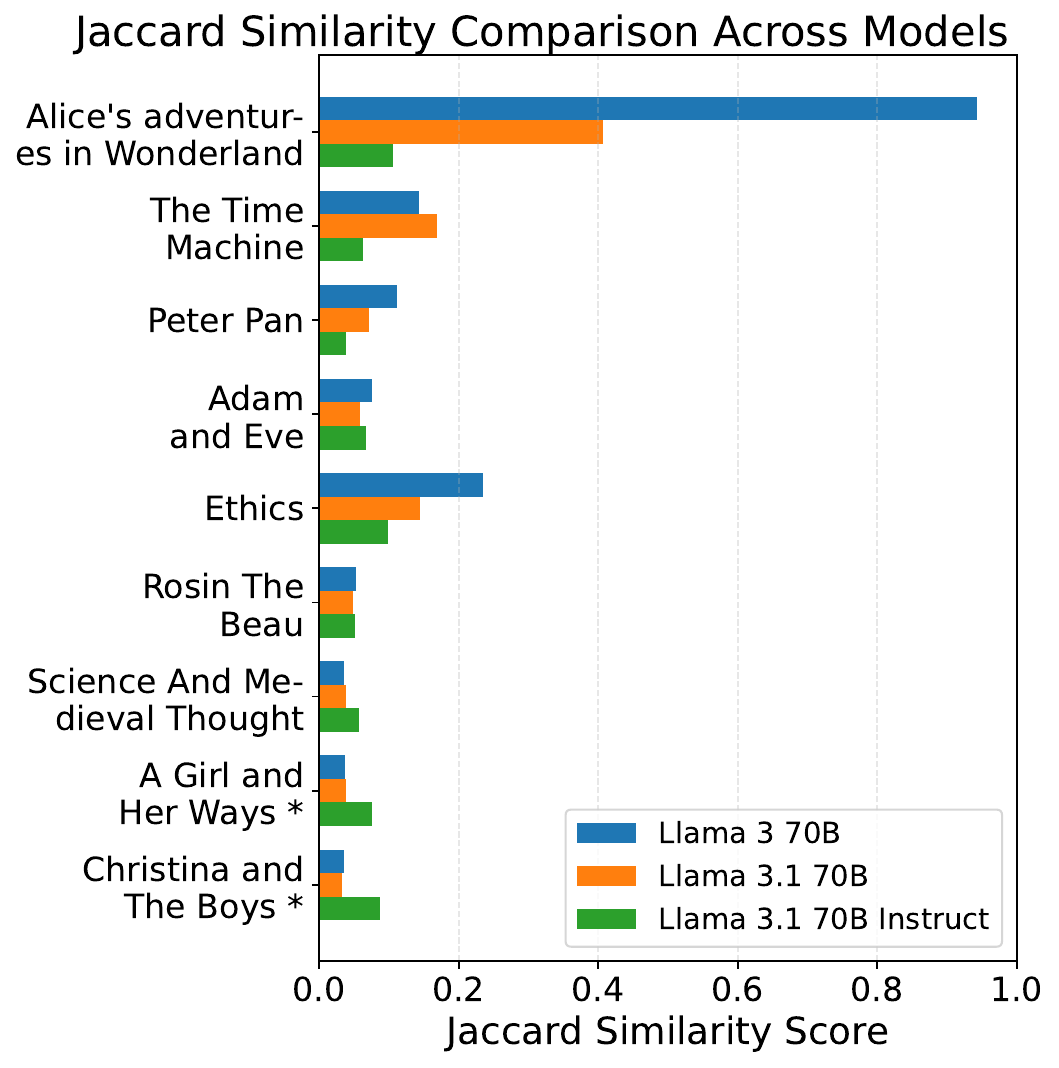}
    \caption{Jaccard Similarity across books for autoregressive generation. * denotes books are added to the Gutenberg after December 2023.}
    \label{fig:ar_jaccard}
\end{figure}

The accompanying Figure~\ref{fig:ar_jaccard}, summarizes these results across all books and models for Jaccard similarity. Figures~\ref{fig:ar_bleu_appendix} and~\ref{fig:ar_rouge_appendix} displays results for BLEU and ROUGE-L similarity, respectively. Each bar represents the model's entire generated output's similarity to the whole ground truth text. Notably, the two post-2023 books are marked with an asterisk to highlight their addition after the Llama 3.1 knowledge cutoff date.

We find that Llama 3 70B exhibits the strongest memorization behavior. It achieves the highest similarity scores on widely popular texts such as \textit{Alice in Wonderland} and \textit{The Time Machine}, supporting the hypothesis that more popular books—likely duplicated across training corpora—are more easily regurgitated by less aligned models.

Llama 3.1 70B typically performs in the middle, showing reduced but still substantial memorization. This suggests that architectural improvements and possible changes to training objectives in Llama 3.1 suppress verbatim memorization while still allowing some training signal retention for popular books.

A particularly interesting counter trend arises with the two books added to Project Gutenberg after Llama 3's training cutoff, denoted with an * in the figure. While both Llama 3 and Llama 3.1 exhibit minimal similarity on these texts, Llama 3.1 Instruct outperforms both, achieving the highest similarity scores across all three metrics. This reversal suggests that instruction tuning, while generally suppressing memorization, may amplify exposure to newer data or surface memorized artifacts.

\subsubsection{Chunk Statistics}

\begin{figure}[t]
      \includegraphics[width=\columnwidth]{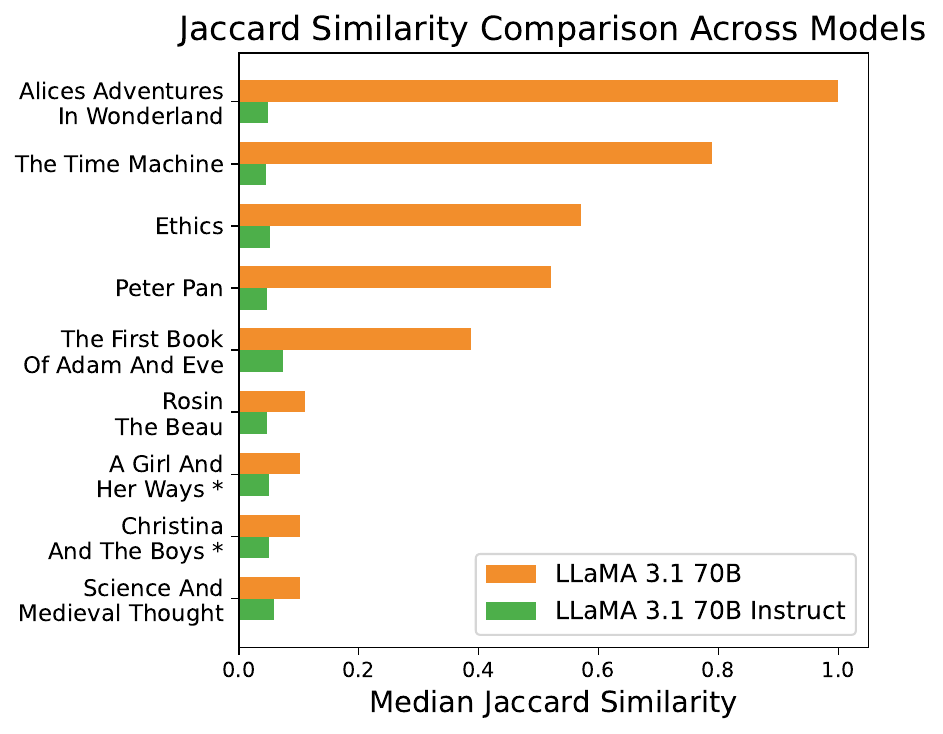}
  \caption{Median Jaccard similarity scores for passage-wise generation on Llama3.1 70B pretrained and Llama3.1 70B Instruct models. * denotes books are added to the Gutenberg after December 2023.}
  \label{fig:01-baseline-models}
\end{figure}

Figure~\ref{fig:01-baseline-models} presents the median Jaccard similarity scores for the nine books for both pretrained  and instruct Llama3.1 70B models. These results were obtained by running prompts for all chunks of the books, without auto-regression.

Extractions for the pretrained  Llama 3.1 demonstrate a noticeably high similarity score ($>$ 0.4) for five books and low score ($<$ 0.2) for four books. \textit{Alice's Adventures in Wonderland} stands out with perfect similarity. Within the four books with low scores, two of them (flagged with an * in the figure) were added to Gutenberg Project after Llama's cutoff date. The low scores of the other two (i.e. \textit{Rosin the Beau} and \textit{Science and Medieval Thought}) can be explained either by their low popularity (see Table~\ref{03-small-data-extraction}) or by their absence from Llama's training data, or both -- we cannot tell.

In contrast, the instruct version of the Llama 3.1 yields uniformly low similarity scores across all nine books, with no single book showing meaningful extraction rates. This strongly indicates that the alignment process significantly reduces the model's direct recall capabilities of specific training data.

\subsection{Exp 2: Supervised Fine-Tuning}

Llama's instruction-tuned models are trained with several mitigations, including some for avoiding verbatim regurgitation of training data. Additional supervised fine-tuning can nudge model to adopt new desired behavior which we can use for data extraction. We finetune Llama3.1 70B and Llama 3.1 70B instruct on the same dataset with two sample sizes: 500 and 1000. 

\begin{figure}[t]
      \includegraphics[width=\columnwidth]{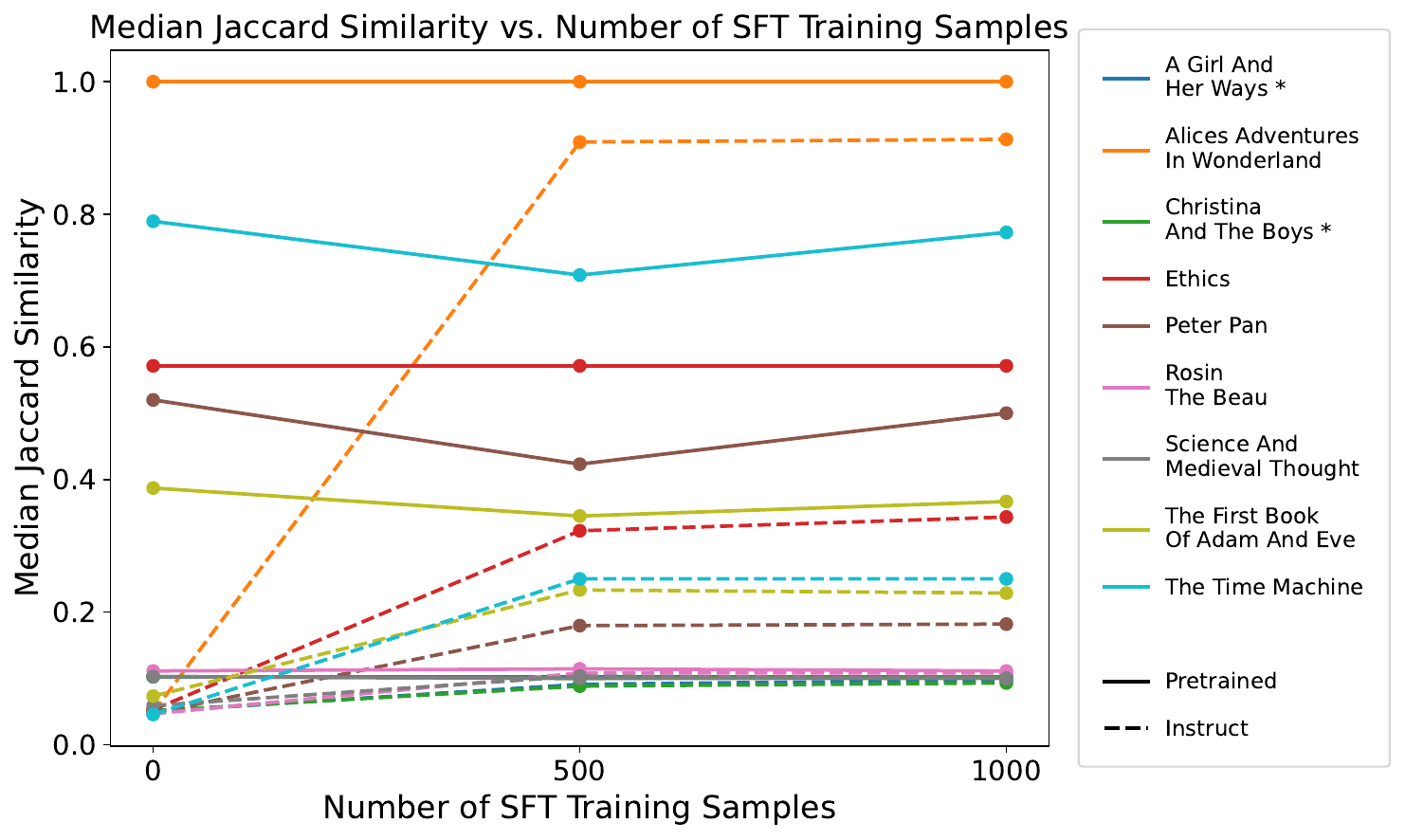}
  \caption{Median Jaccard scores for fine-tuned models (pretrained  \& instruct) on different sample sizes. * denotes books are added to the Gutenberg after December 2023.}
  \label{fig:02-supervised-fine-tune}
\end{figure}

Figure~\ref{fig:02-supervised-fine-tune} presents the impact of fine-tuning on data extraction performance for both pretrained (solid lines) and instruct (dashed lines) variants of the Llama 3.1 70B model. Fine-tuning of the pretrained model (solid lines) does not seem to affect much the extraction rates with respect to the baseline of $x=0$, as seen by the mostly-horizontal lines throughout. If anything, it slightly disturbs the performance, at least until there are enough samples (1,000) to reinforce the recall task. 

Extraction rates drop significantly in the instruction-tuned model (dashed lines) without additional fine-tuning ($x=0$), being nearly noise for all the books. However, after fine-tuning with 500 samples or more, the similarity scores for five of the books increase noticeably. Here, too, \textit{Alice's Adventures in Wonderland} stands out, with an extraction rate around 90\%, closely matching the pretrained baseline. The extraction rates for four of the books do not seem to improve with SFT. These are the same books discussed before, two of which were added after Llama's cutoff date and two that are largely unknown and/or may not have been in the training data.

These results are along the lines of those in Nasr et al.~\cite{28nasr2025scalable}, and show that instruction tuning primarily alters how the model interacts with users, rather than significantly affecting its internal memorization of training data.

\subsection{Exp 3: Llama 3.1-instruct SFT-1000}




To further investigate the influence of popularity on memorization performance at scale, we expanded our experiments by using the Llama3.1-Instruct fine-tuned with 1,000 samples. Specifically, we evaluated the model's memorization across an expanded set of 32 books. Results from this expanded experiment are presented in Figure~\ref{fig:03-sft-1000}.

As shown in the figure, books with higher number of ratings generally achieve significantly higher median Jaccard similarity scores compared to books with lower number of ratings. The correlation coefficient is 0.5, which is indicative of a fairly strong positive correlation. This correlation suggests that higher popularity may be associated with greater availability and duplication on the internet. The three books with the highest reconstruction rates are \textit{The Communist Manifesto} (0.95), \textit{Alice's Adventures in Wonderland} (0.91), and \textit{Romeo and Juliet} (0.76). See Appendix~\ref{sec:additional-results} for more details.

With respect to the books added after the cutoff date (red dots), their popularity does not seem to change the extraction rate, meaning that, with very high likelihood, none of these books were in the training data of Llama 3.


Overall, the expanded fine-tuning experiment confirms the important role of popularity, possibly as a proxy of duplication, in determining extraction rate using this ``prefix-prompting'' extraction method.


\section{Analysis of Weight Updates}

In this section, we focus on analyzing the weight updates introduced by the LoRA fine-tuning process on the baseline Llama model. By design, LoRA only updates certain layers of the original network. Moreover, due to the compressed nature of the LoRA formalism — where the rank of the learned adaptation matrices is usually much smaller than the full rank of the underlying weight matrices — only a subset of the parameters within those layers are effectively modified. This raises two natural questions: how large is the fraction of the original weights that receive significant updates, and how are these updates distributed across the different layers of the network?

To address these questions, we focus on our SFT-1000 trained LoRA model as a representative case. Our trained LoRA models were free to modify layers of the following modules of the Llama transformer blocks (q, k, v and o), at the self-attention and feedforward MLP blocks (gate, up, down). Figure \ref{fig:llama-lora-block} provides a schematic overview of the Llama transformer architecture, highlighting the locations where our LoRA adapters are integrated.

To carry out our analysis, we begin by reconstructing the weight update matrices for all layers that could be modified by the LoRA adapters. We note that the Llama 3.1 70B Instruct model has 80 stacked transformer blocks containing all the aforementioned modules. Thus, since our LoRA training was applied across the full model depth, we reconstruct 560 weight update matrices. As typical in LoRA, each of the layers we decided to train adaptors gets a pair of low-rank matrices $A$ and $B$ of shapes $A \in \mathbb{R}^{r \times d_{\text{in}}}$ and $B \in \mathbb{R}^{d_{\text{out}} \times r}$, where $d_{\text{in}}$ and $d_{\text{out}}$ are the input and output dimensions of the original Llama weight matrix $W \in \mathbb{R}^{d_{\text{out}} \times d_{\text{in}}}$ for that layer. The rank we use for our LoRA adaptors is $r = 16$.

To study where significant updates take place, we must perform the reconstruction of the weight update $W_{\mathrm{update}}$ matrix from LoRA's $A$ and $B$ trained matrices. This reconstruction is straightforward: the full-rank weight update is simply $ W_{\text{update}} = \alpha r^{-1} \cdot B A $, where $\alpha$ is the LoRA scaling factor and $r$ is the LoRA rank hyperparameter. This update matrix has the same shape as the original weight matrix, i.e. $W_{\text{update}} \in \mathbb{R}^{d_{\text{out}} \times d_{\text{in}}}$, and represents the effective change that would be applied to the base model if the LoRA adapters were merged back into the baseline Llama model. 

Since LoRA's central idea is training these low-rank projections while keeping the original model weights frozen, $W_{\text{update}}$ captures exactly what LoRA is trying to inject into the base model after the supervised fine-tuning process, which is precisely what we want to discover. Nevertheless, it is misleading to analyse $W_{\text{update}}$ directly, since what really matters is the impact of the update in the original network weights, and not the absolute values of these updates: a small absolute $\Delta$ value of the update might actually cause a huge impact if the original neuron weight was tiny, while a large $\Delta$ might be insignificant if the original weight was already huge. Thus, we further construct relative update matrices, i.e.,:

\[
W_{\mathrm{rel}} = \mathrm{abs}(W_\mathrm{update} \oslash W_\mathrm{original})
\]

\noindent where $\oslash$ denotes a Hadamard division (which is just an element-wise division for matrices of equal dimensions, as here), and $\mathrm{abs}$ denotes that the matrix has all its elements in absolute, positive values.

\begin{figure}
    \centering
    \includegraphics[width=1\linewidth]{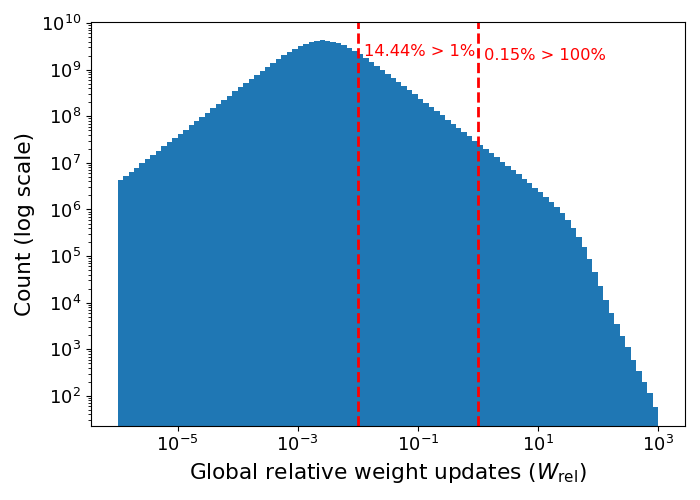}
    \caption{Log-log scale histogram of relative updates of individual weights in the entire network. The vast majority of updates are relatively small compared to the original Llama weights.}
    \label{fig:global-updates}
\end{figure}

Figure \ref{fig:global-updates} shows a histogram built from the concatenated set of values of all 560 $W_{\mathrm{rel}}$ matrices. The distribution clearly reveals that the vast majority of updates are relatively small in magnitude when compared to the original weights. Only  about $\sim14\%$ of the original weights are receiving a boost greater than only $1\%$, and a mere $\sim0.15\%$ are updated by more than 100\%. These results suggest that only sparse and highly localized updates are sufficient to make the instruct network start remembering documents that were used in its training set.

\begin{figure}
    \centering
    \includegraphics[width=0.8\linewidth, trim=0 0 0 0, clip]{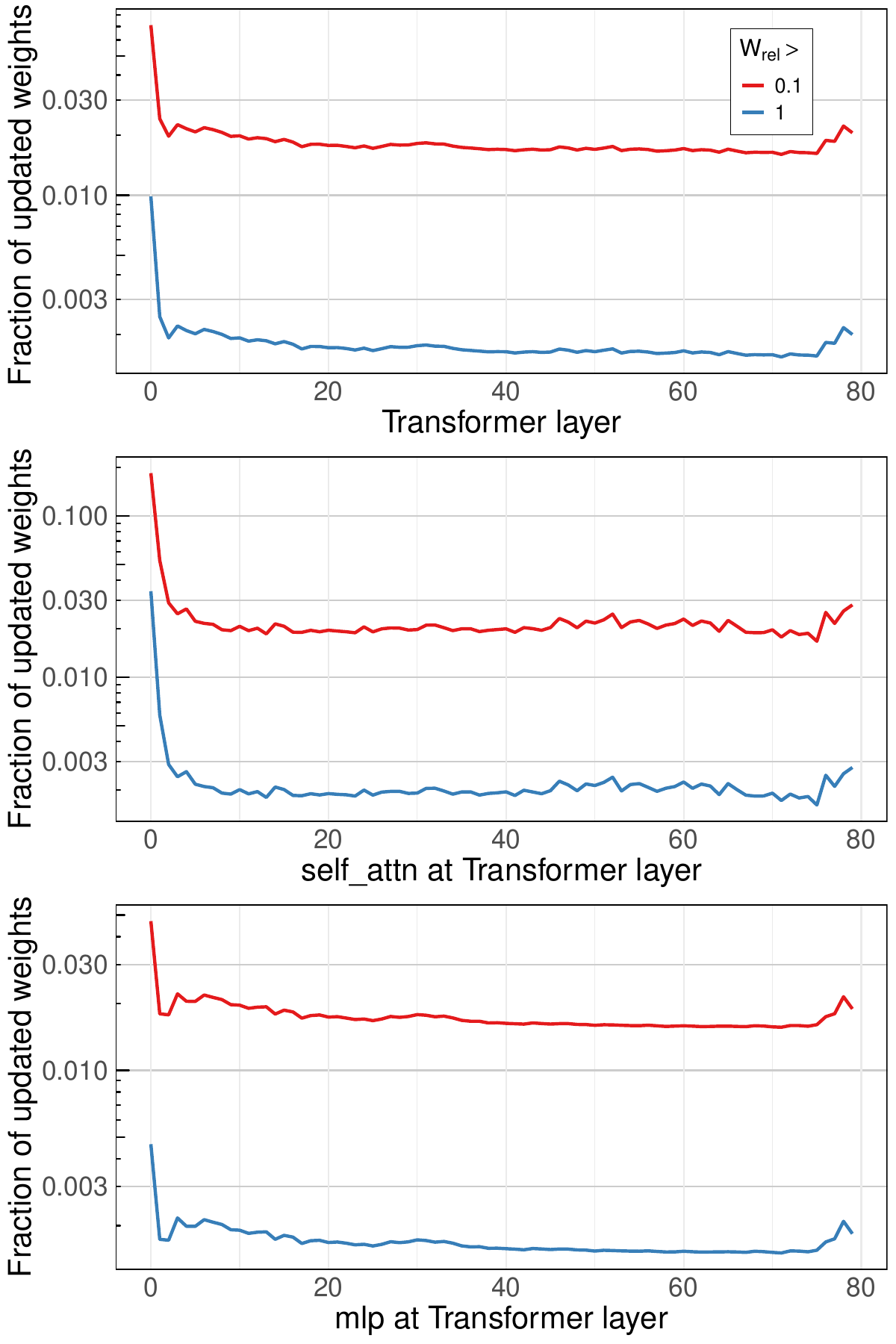}
    \caption{Fraction of updated weights across the transformer layers of Llama-3.1 after the SFT-1000 LoRA fine-tuning. The top panel shows the fractions for the entire transformers; the middle panel for all self-attention layers; and the bottom panel for all MLP layers. The curves indicate the fraction of weights whose relative update magnitude exceeds the thresholds of $W_{\text{rel}} > 0.1$ (red curves) and $W_{\text{rel}} > 1$ (blue curves).}
    \vspace{-18pt}
    \label{fig:updatesPerlayer}
\end{figure}

Naturally, this result raises the follow-up question of how these few significant updates are distributed across the entire Llama network. The top panel of Figure \ref{fig:updatesPerlayer} shows that these updates are heavily concentrated at the  earliest transformers instead of the significant updates being applied more uniformly throughout the entire network. This pattern is similar regardless of whether we examine the self-attention blocks or the multilayer perceptron (MLP) blocks. Nevertheless, although the evolution of the update fraction along the network is similar for both types of blocks -- with a predominant concentration of significant updates in the early layers -- we observe that the self-attention layers receive approximately seven times more updates (in fraction) than the MLP layers in these early transformers. These findings suggest that early layers play a central role in adapting the model, while later layers require minimal changes to help the network remember its training data. Moreover, even at those early layers, the fraction of weights of the original network that needs to be changed is sparse.

\section{Conclusion}
\label{sec:conclusion}

Our results demonstrate that modern large language models, particularly the Llama 3 family, retain substantial amounts of memorized content from the training corpus. We find that both autoregressive generation, and passage-wise reconstruction are sensitive to a book's likely presence in the training data, with stronger reconstruction observed for more popular or widely reviewed texts.

Instruction tuned models like Llama 3.1 exhibit reduced memorization by default, but we show that targeted fine-tuning can partially mitigate this suppression. This effect is most pronounced in the lower layers of the network, where small updates appear to undo alignment caused suppression.

More broadly, our study introduces a scalable framework for measuring memorization across models and training stages. By combining behavioral evaluations with an analysis in the change of the weights of these models, we uncover correlations between memorization, training exposure, and popularity, offering insight into what factors make a model remember, and when that memory can be accesed or suppressed.

\section{Limitations}
\label{sec:limitations}

Our study provides initial evidence on the extent of book extraction from LLMs, but several limitations should be noted, which in turn suggest directions for future research.

First, this study is limited to the Llama 3.x family of models, specifically the 70B parameter variants. While this focus enables a detailed examination of memorization and extraction within a widely used model, it restricts the generalizability of our findings to other model families and architectures. We did not investigate scaling effects or compare across different model sizes, as prior work ~\cite{01carlini2023quantifying} has consistently demonstrated that memorization capacity increases with model size.

Second, our evaluation primarily targets books that are publicly available, particularly those released on Project Gutenberg prior to the Llama 3.x knowledge cutoff date. Future work could extend this analysis to books released after the cutoff date, as well as to widely known but copyright-protected works that are not part of the Project Gutenberg collection, such as \textit{A Farewell to Arms} and the \textit{Harry Potter} series.

Finally, we use book popularity, as measured by Goodreads ratings, as a proxy for the likelihood of duplication in the training data. While this approach is practical for published books, it may not generalize to other types of content, such as news articles or academic papers. For these domains, alternative metrics (e.g., number of downloads or citations) may be required, but their suitability as proxies for training data frequency remains to be validated.

\bibliography{main}

\appendix

\section{Book Details}
\label{sec:book-details}

The following tables provide detailed information for all books used in this study, including their titles, number of Goodreads ratings, authors, and initial Project Gutenberg release dates. The books are sorted in descending order of popularity. Table~\ref{tab:book_train_details} lists the 43 books used for fine-tuning, while Table~\ref{tab:book_testing_details} lists the 32 books used for testing.

\begin{table*}[]
    \centering
    \resizebox{1\textwidth}{!}{%
    \begin{tabular}{llll}
        \toprule
        \textbf{Ratings} & \textbf{Book Names} & \textbf{Author} & \textbf{First Added On} \\
        \midrule
        5630463 & The Great Gatsby & F. Scott Fitzgerald & January 17, 2021 \\
        4539830 & Pride and Prejudice & Jane Austen & June 1, 1998 \\
        2199001 & Jane Eyre: An Autobiography & Charlotte Brontë & March 1, 1998 \\
        1929915 & Wuthering Heights & Emily Brontë & December 1, 1996 \\
        1307593 & Adventures of Huckleberry Finn & Mark Twain & June 29, 2004 \\
        1293875 & Metamorphosis & Franz Kafka & August 17, 2005 \\
        1250187 & Sense and Sensibility & Jane Austen & September 1, 1994 \\
        1141312 & The Odyssey & Homer & April 1, 1999 \\
        999417 & Crime and Punishment & Fyodor Dostoyevsky & March 28, 2006 \\
        988702 & A Tale of Two Cities & Charles Dickens & January 1, 1994 \\
        984922 & The Adventures of Tom Sawyer, Complete & Mark Twain & July 1, 2004 \\
        975420 & The Count of Monte Cristo & Alexandre Dumas, Auguste Maquet & January 1, 1998 \\
        886396 & A Christmas Carol in Prose; Being a Ghost Story of Christmas & Charles Dickens & August 11, 2004 \\
        848348 & Great Expectations & Charles Dickens & July 1, 1998 \\
        742250 & Persuasion & Jane Austen & February 1, 1994 \\
        631406 & The Strange Case of Dr. Jekyll and Mr. Hyde & Robert Louis Stevenson & June 27, 2008 \\
        539036 & Heart of Darkness & Joseph Conrad & January 9, 2006 \\
        485278 & The Iliad & Homer & July 1, 2004 \\
        392898 & The Importance of Being Earnest: A Trivial Comedy for Serious People & Oscar Wilde & March 1, 1997 \\
        364727 & The Prince & Niccolò Machiavelli & February 11, 2006 \\
        361107 & The Brothers Karamazov & Fyodor Dostoyevsky & February 12, 2009 \\
        352180 & War and Peace & graf Leo Tolstoy & April 1, 2001 \\
        328677 & An Anglo-Saxon Epic Poem & J. Lesslie Hall (translator) & July 19, 2005 \\
        328025 & The Yellow Wallpaper & Charlotte Perkins Gilman & November 1, 1999 \\
        317240 & The Adventures of Sherlock Holmes & Arthur Conan Doyle & March 1, 1999 \\
        217766 & Grimms' Fairy Tales & Jacob Grimm, Wilhelm Grimm & April 1, 2001 \\
        217052 & The Republic & Plato & October 1, 1998 \\
        166045 & Thus Spake Zarathustra: A Book for All and None & Friedrich Wilhelm Nietzsche & December 1, 1999 \\
        136898 & Ulysses & James Joyce & July 1, 2003 \\
        130667 & Narrative of the Life of Frederick Douglass, an American Slave & Frederick Douglass & January 12, 2006 \\
        106656 & Beyond Good and Evil & Friedrich Wilhelm Nietzsche & August 1, 2003 \\
        69676 & The Confessions of St. Augustine & Bishop of Hippo Saint Augustine & June 1, 2002 \\
        50681 & Leviathan & Thomas Hobbes & May 1, 2002 \\
        48030 & A Modest Proposal & Jonathan Swift & October 1, 1997 \\
        46229 & Cranford & Elizabeth Cleghorn Gaskell & January 1, 1996 \\
        43921 & The Souls of Black Folk & W. E. B. Du Bois & January 1, 1996 \\
        38189 & Walden, and On The Duty Of Civil Disobedience & Henry David Thoreau & January 1, 1995 \\
        23339 & Second Treatise of Government & John Locke & January 1, 2005 \\
        21501 & The King in Yellow & Robert W. Chambers & July 1, 2005 \\
        2726 & The Letters of Jane Austen & Jane Austen & February 12, 2013 \\
        2548 & The Works of Edgar Allan Poe — Volume 2 & Edgar Allan Poe & April 1, 2000 \\
        1277 & The Adventures of Roderick Random & T. Smollett & May 1, 2003 \\
        65 & The Devil is an Ass & Ben Jonson & October 7, 2015 \\
        \bottomrule
    \end{tabular}%
    }
    \caption{The fine-tuning dataset, consisting of 43 books from Project Gutenberg released before December 2023, sorted by the number of Goodreads ratings.}
    \label{tab:book_train_details}
\end{table*}
\begin{table*}
    \centering
    \resizebox{1\textwidth}{!}{%
    \begin{tabular}{llll}
        \toprule
        \textbf{Ratings} & \textbf{Book Names} & \textbf{Author} & \textbf{First Added On} \\
        \midrule
        2,735,023 & Romeo and Juliet & William Shakespeare & November 1, 1998 \\
        831,152 & Siddhartha & Hermann Hesse & February 1, 2001 \\
        546,286 & The Time Machine & H. G. Wells & October 2, 2004 \\
        492,129 & The Wonderful Wizard of Oz & L. Frank Baum & October 12, 2013 \\
        413,400 & Alice's Adventures in Wonderland & Lewis Carroll & June 27, 2008 \\
        362,694 & Peter Pan & J. M. Barrie & June 25, 2008 \\
        290,524 & Candide & Voltaire & November 27, 2006 \\
        219,332 & The Tempest & William Shakespeare & October 26, 2007 \\
        198,834 & Notes from the Underground & Fyodor Dostoyevsky & July 1, 1996 \\
        183,818 & The Communist Manifesto & Karl Marx, Friedrich Engels & January 25, 2005 \\
        169,009 & The Sign of the Four & Arthur Conan Doyle & March 1, 2000 \\
        147,072 & The Legend of Sleepy Hollow & Washington Irving & June 27, 2008 \\
        140,217 & Through the Looking-Glass & Lewis Carroll & June 25, 2008 \\
        129,512 & The Island of Doctor Moreau & H. G. Wells & October 14, 2004 \\
        48,922 & Just So Stories & Rudyard Kipling & August 1, 2001 \\
        19,734 & Ethics & Benedictus de Spinoza & February 1, 2003 \\
        7,582 & The Aesop for Children & Aesop & December 2, 2006 \\
        6,071 & The Secret of the Caves* & Franklin W. Dixon & February 7, 2025 \\
        787 & Simple Sabotage Field Manual & United States. Office of Strategic Services & August 4, 2008 \\
        344 & The First Book of Adam and Eve & Rutherford Hayes Platt & January 1, 1996 \\
        138 & The Philippines a Century Hence & Austin Craig & April 18, 2011 \\
        22 & The Emma Gees & Herbert W. McBride & February 24, 2007 \\
        16 & Bab Ballads and Savoy Songs & W. S. Gilbert & March 15, 2005 \\
        5 & Dragon Moon* & Henry Kuttner & January 28, 2025 \\
        4 & The Hallowell Partnership & Katharine Holland Brown & October 14, 2012 \\
        2 & Judas Ram & Sam Merwin & January 27, 2016 \\
        2 & Rosin the Beau & Laura Elizabeth Howe Richards & December 24, 2008 \\
        0 & Christina and the Boys* & Amy Le Feuvre & February 10, 2025 \\
        0 & Pegasus* & J. F. C. Fuller & January 30, 2025 \\
        0 & A girl and her ways* & Amy Le Feuvre & January 28, 2025 \\
        0 & Upside Down or Backwards & W. C. Tuttle & December 20, 2021 \\
        0 & Science and Medieval Thought & T. Clifford Allbutt & February 21, 2012 \\

        \bottomrule
    \end{tabular}%
    }
    \caption{The testing dataset, consisting of 32 books released both before and after the cutoff date, sorted by the number of Goodreads ratings. * denotes a book released \textbf{after} the knowledge cutoff date (December 2023).}
    \label{tab:book_testing_details}
\end{table*}
\section{Additional Results}
\label{sec:additional-results}

Here, we include experimental results that are not displayed, but support and extend the claims and findings in the main body of the paper.

\subsection{Autoregressive Generation Experiments}

\begin{figure}
    \centering
        \includegraphics[width=\linewidth]{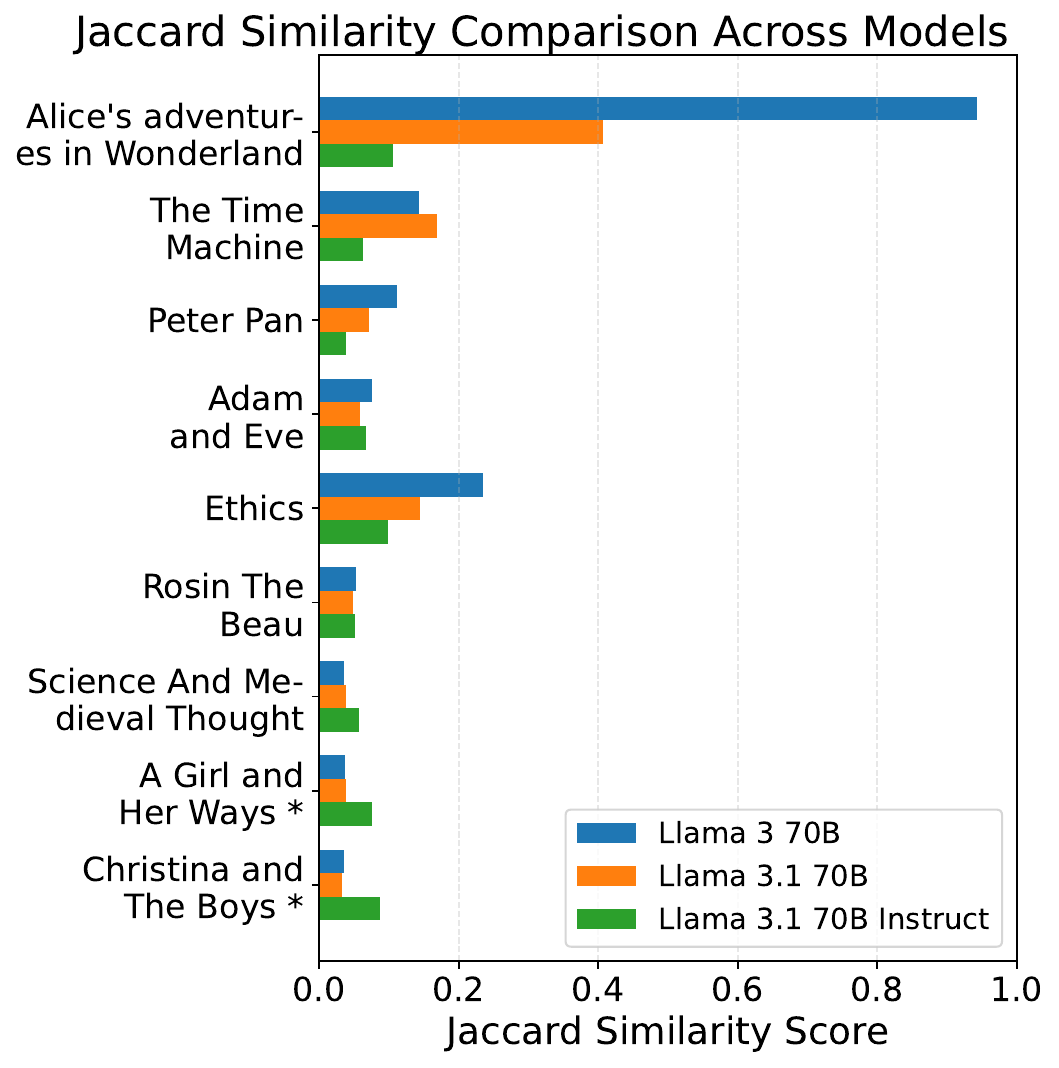}
    \caption{Jaccard Similarity across books for autoregressive generation. * denotes books are added to the Gutenberg after December 2023.}
    \label{fig:ar_jaccard_appendix}
\end{figure}

\begin{figure}
    \centering
    \includegraphics[width=\linewidth]{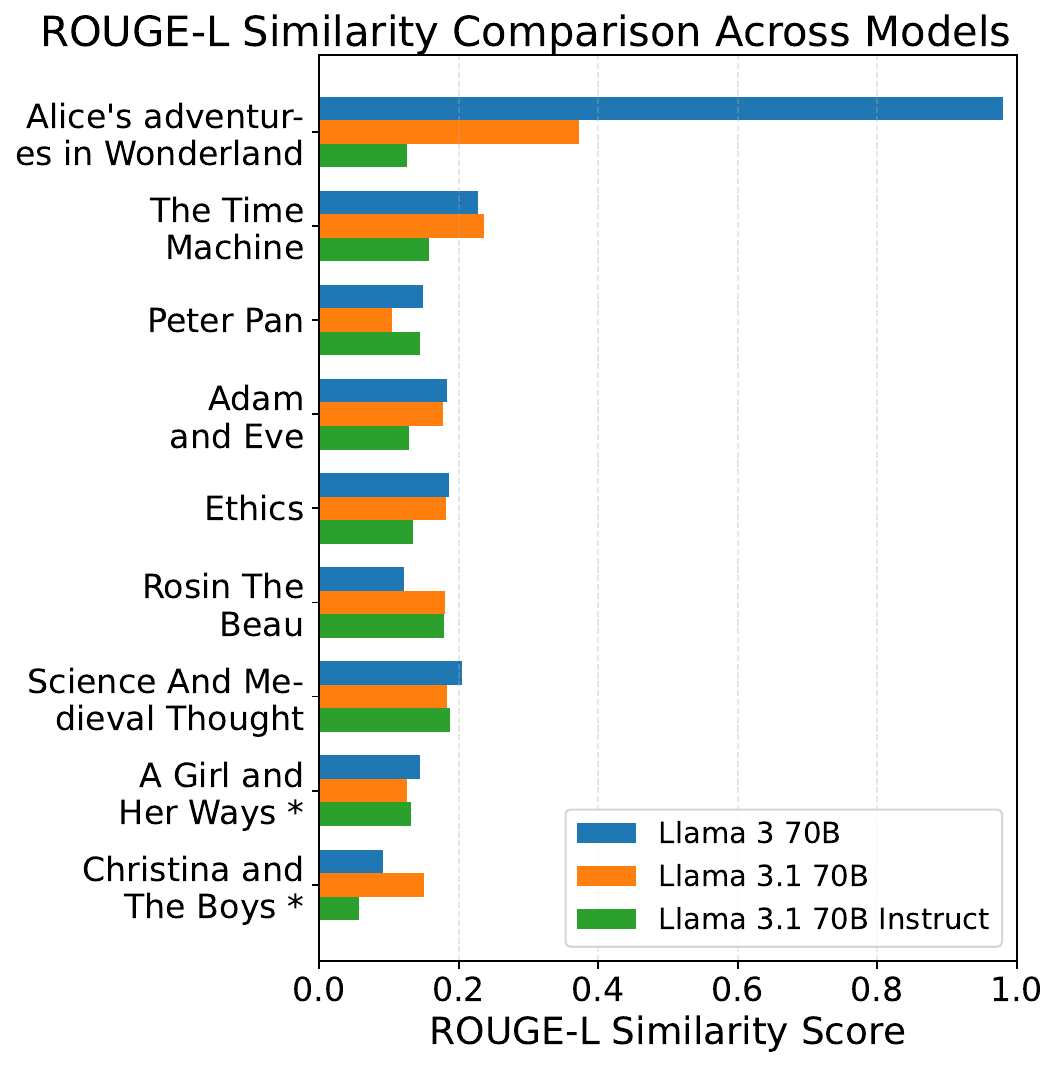}
    \caption{ROUGE-L Score across books for autoregressive generation. * denotes that these books were added to the Gutenberg repository after December 2023}
    \label{fig:ar_rouge_appendix}
\end{figure}

\begin{figure}
    \centering
    \includegraphics[width=\linewidth]{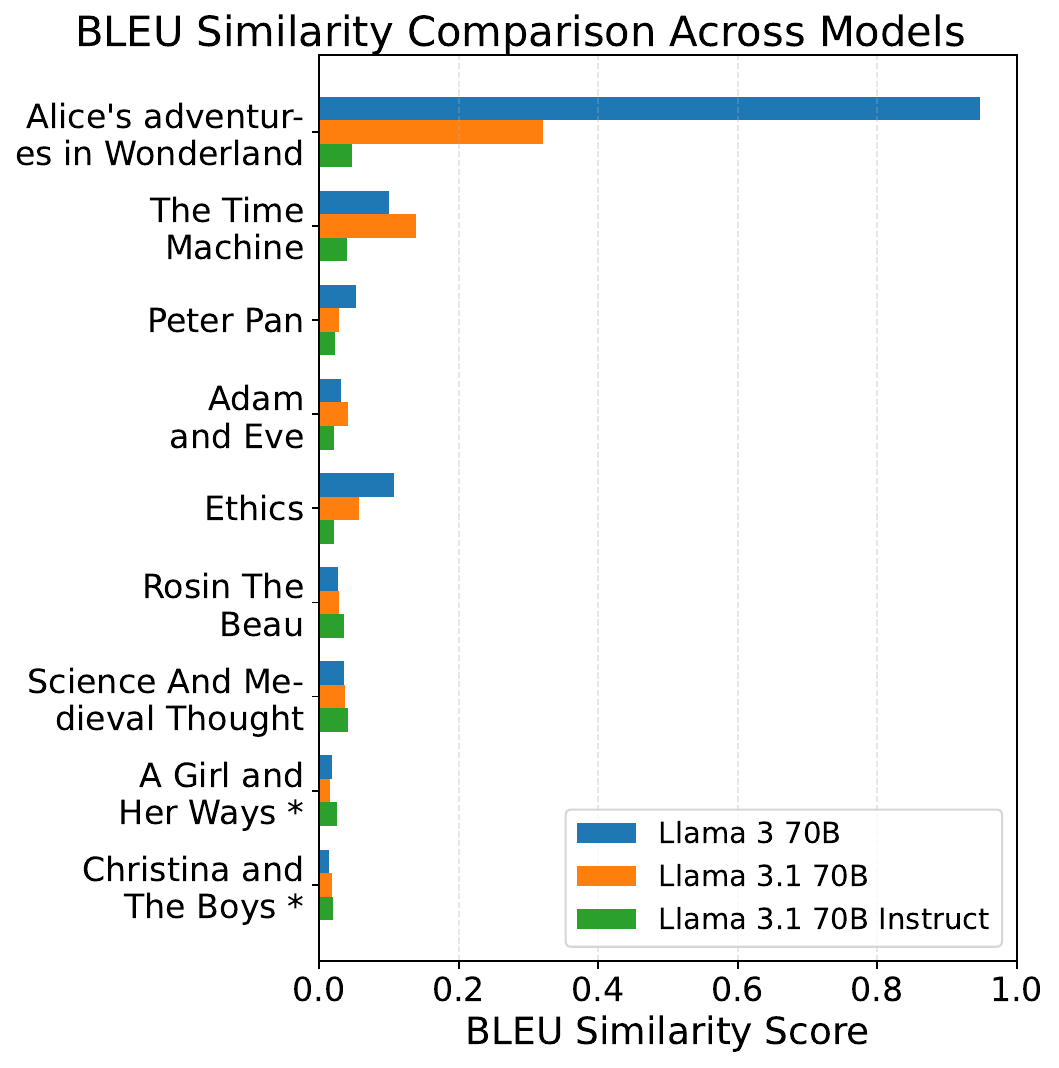}
    \caption{BLEU Score across books for autoregressive generation. * denotes that these books were added to the Gutenberg repository after December 2023}  
    \label{fig:ar_bleu_appendix}
\end{figure}

This section presents the full set of similarity metrics used to evaluate autoregressive generation, including Jaccard ( Figure~\ref{fig:ar_jaccard_appendix}), ROUGE-L (Figure~\ref{fig:ar_rouge_appendix}), and BLEU (Figure~\ref{fig:ar_bleu_appendix}).

\begin{itemize}
    \item Jaccard similarity, which was the focus of the main text, estimates memorization based on exact token overlap over the set of tokens from the original tokens, and those that are autoregressively generated.
    \item ROUGE-L focuses on the longest common subsequence between the text
    \item BLEU measures n-gram precision and is sensitive to shorter patterns
\end{itemize}


Across all three metrics, we observe consistent trends:
\begin{itemize}
    \item Llama 3 70B demonstrates the highest memorization, especially for older, more popular books.
    \item Llama 3.1 70B generally falls in the middle, indicating reduced verbatim recall.
    \item Llama 3.1 instruct performs best on newer books, indicatinf that alignment or intruction tuning may unintentionally enhance memorization of more recent content.
\end{itemize}

\subsection{Llama 3.1-instruct SFT-1000}
\begin{figure}
    \centering
    \includegraphics[width=\linewidth]{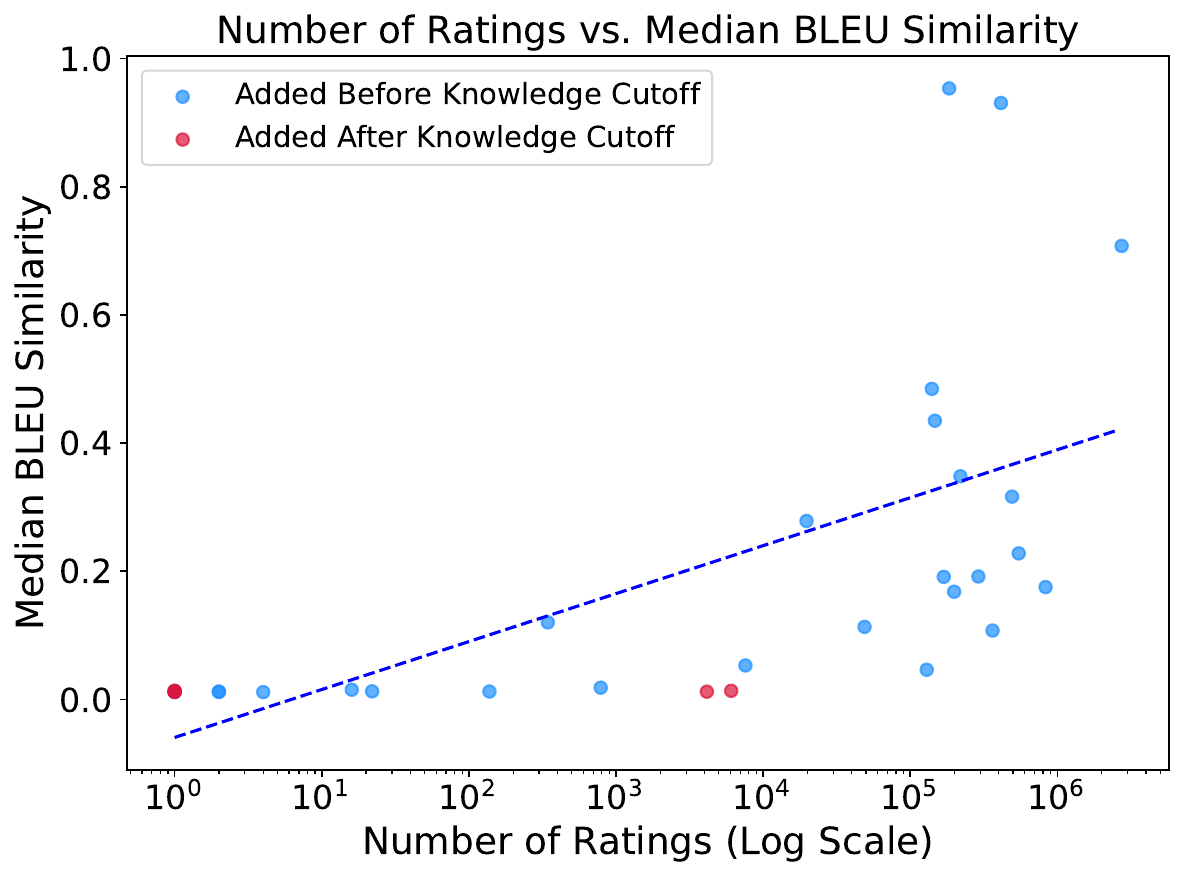}
    \caption{Median BLEU scores}
    \label{fig:exp3-bleu}
\end{figure}

\begin{figure}
    \centering
    \includegraphics[width=\linewidth]{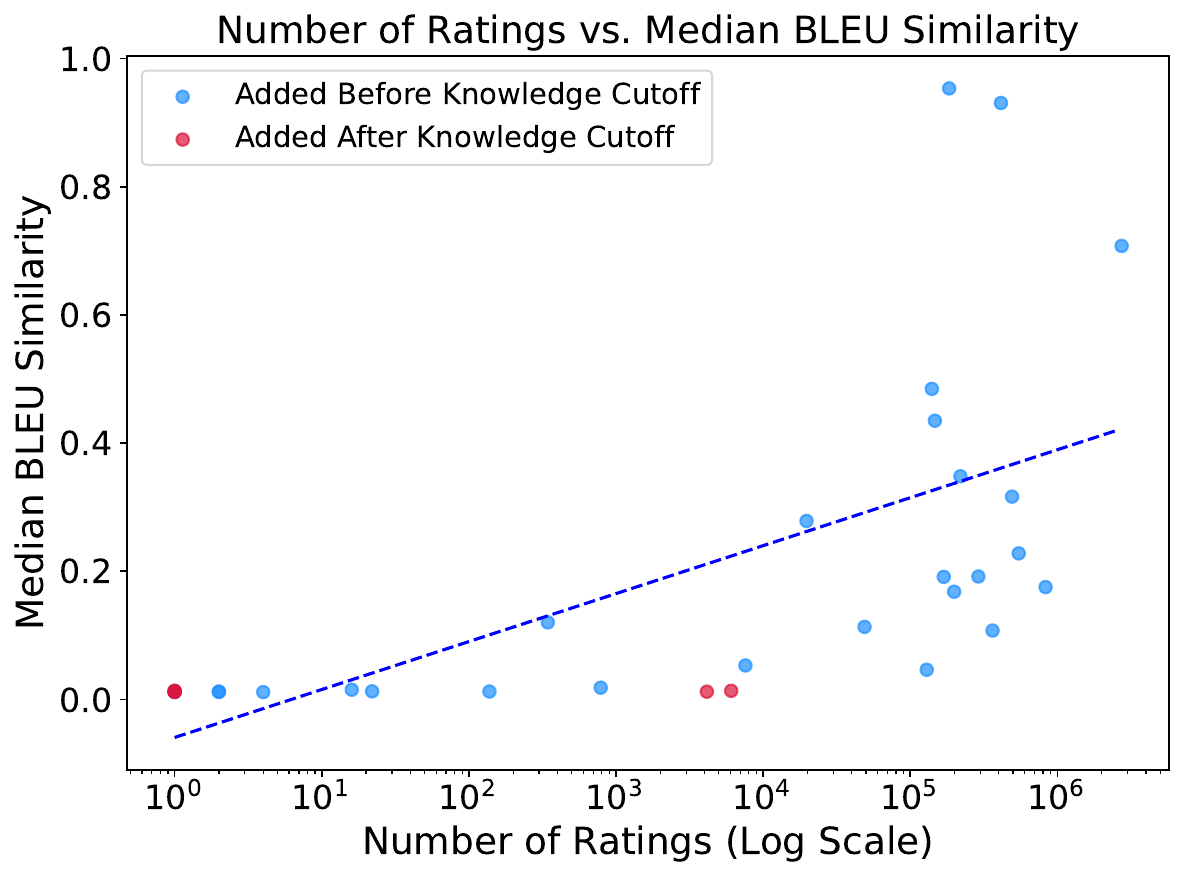}
    \caption{Median Cosine Similarity scores}
    \label{fig:exp3-cosine}
\end{figure}

\begin{figure}
    \centering
    \includegraphics[width=\linewidth]{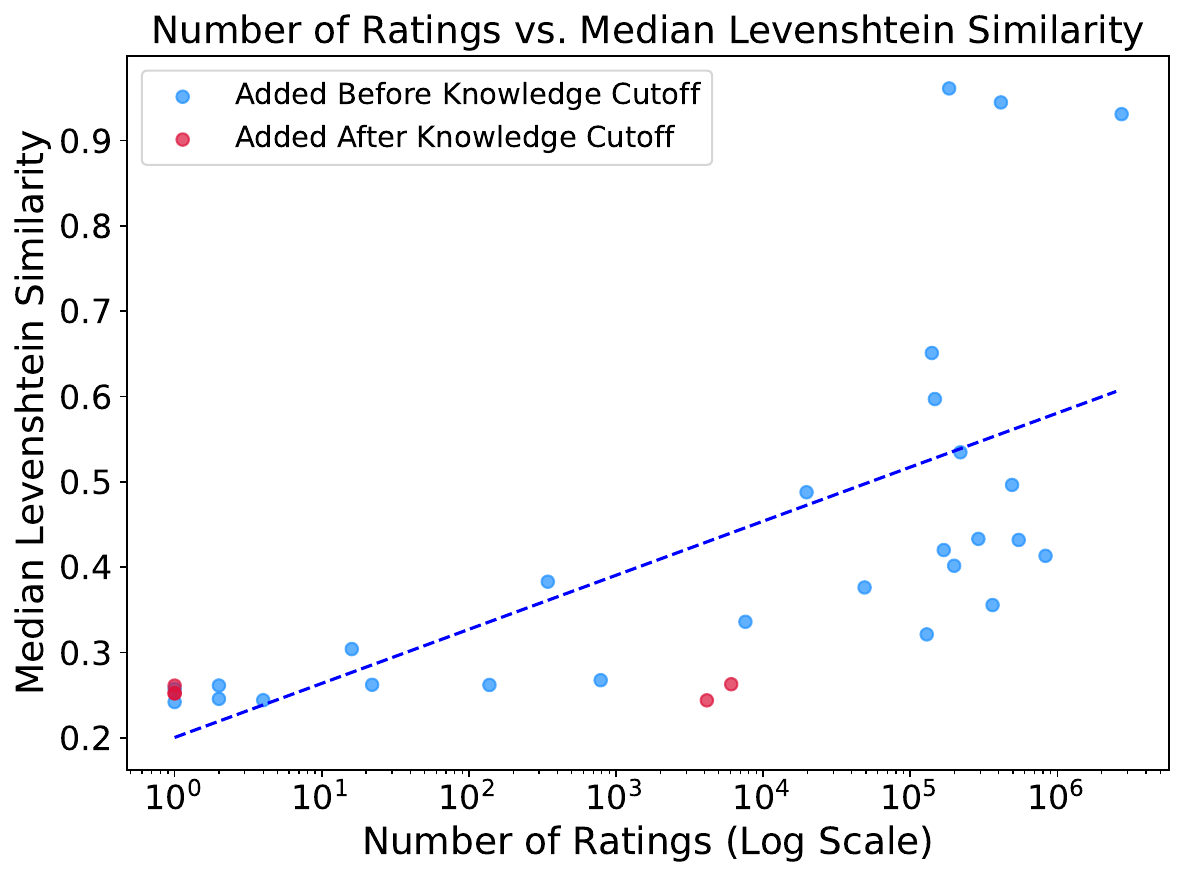}
    \caption{Median Levenshtein scores}
    \label{fig:exp3-levenshtein}
\end{figure}

\begin{figure}
    \centering
    \includegraphics[width=\linewidth]{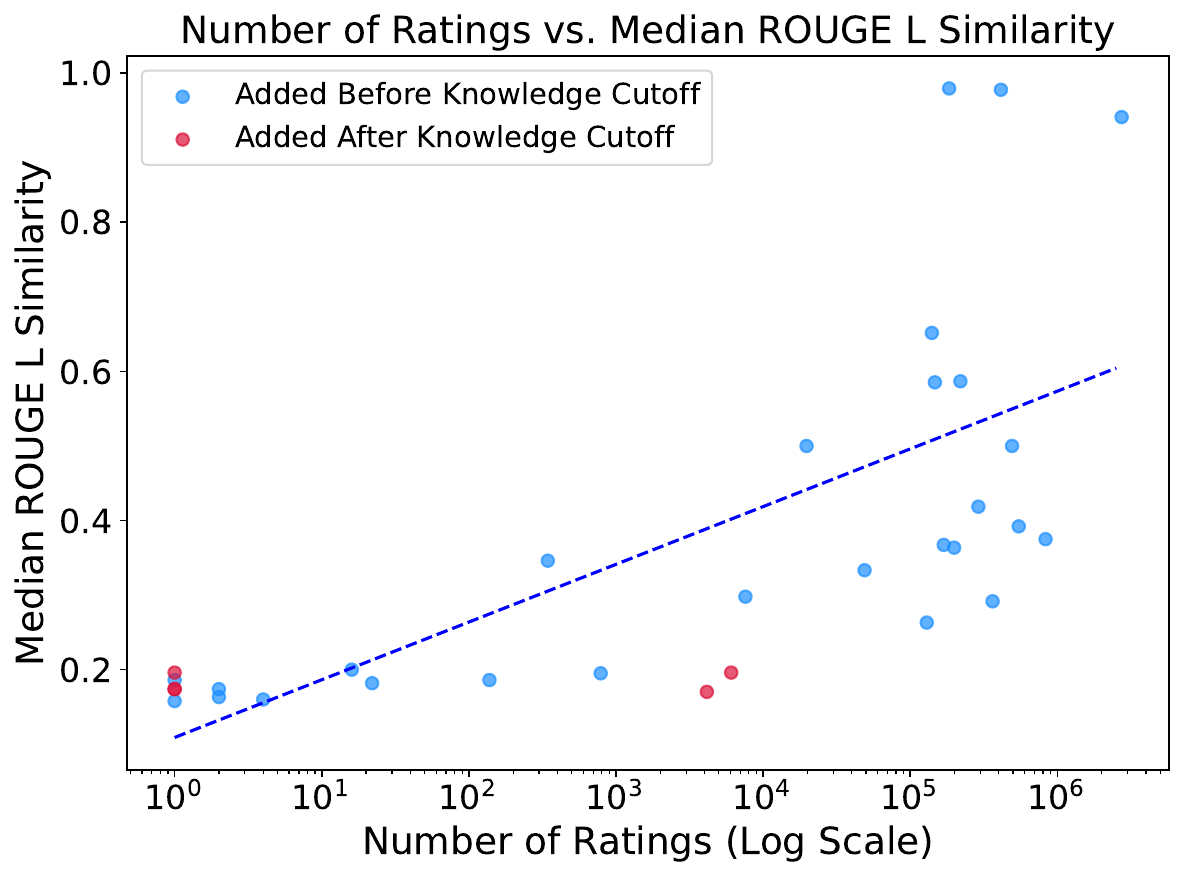}
    \caption{Median ROUGE-L scores}
    \label{fig:exp3-rouge-l}
\end{figure}

\begin{figure}
    \centering
    \includegraphics[width=\linewidth]{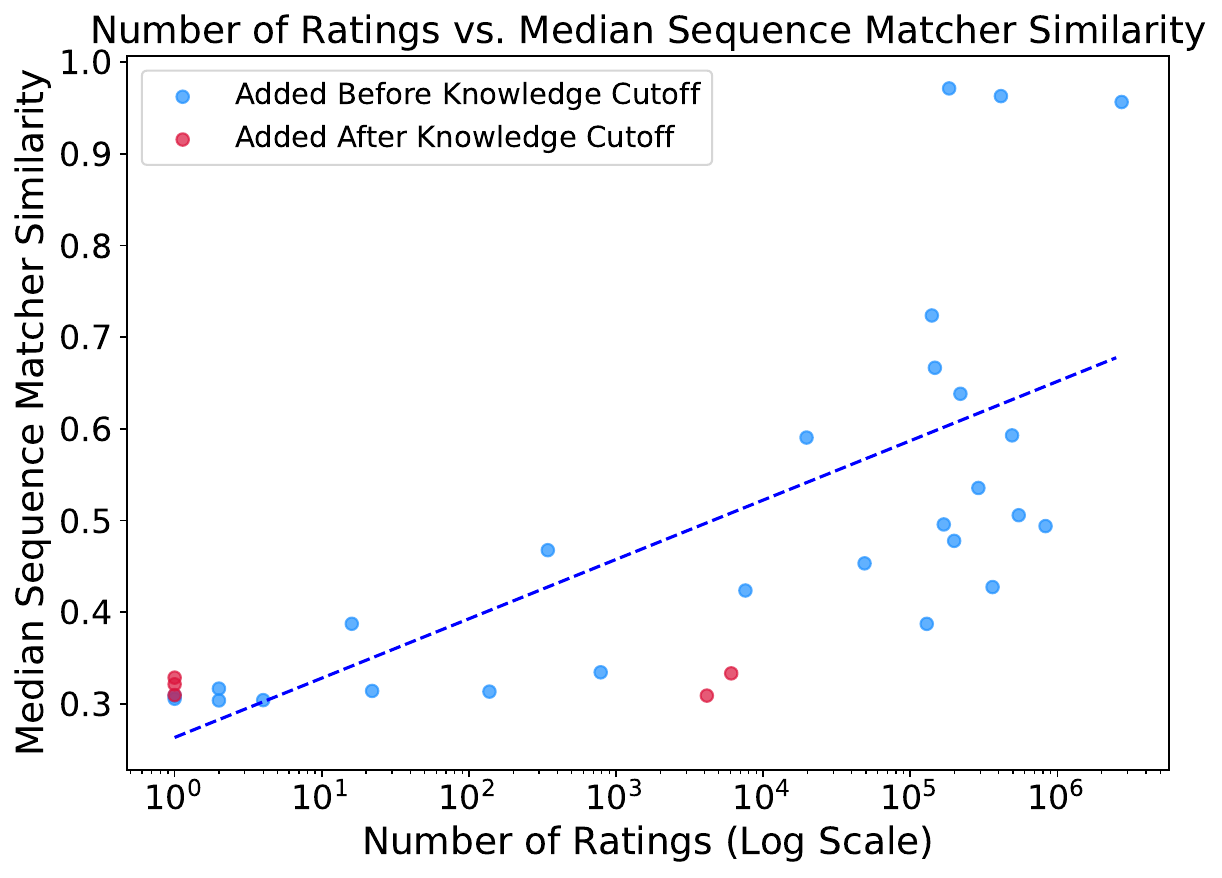}
    \caption{Median Sequence Matcher scores}
    \label{fig:exp3-sequence-matcher}
\end{figure}

Median similarity scores for books extracted from Llama 3.1-instruct SFT-1000 samples are presented. Books from the pre-training collection (\textcolor{blue}{pre}) and post-training collection (\textcolor{red}{post}) are indicated by blue and red markers, respectively. The exact median similarity scores are listed in Table~\ref{tab:book_testing_scores}. Visualizations of these results are shown in Figures~\ref{fig:exp3-bleu}--\ref{fig:exp3-sequence-matcher}.
\begin{table*}
    \centering
    \resizebox{1\textwidth}{!}{%
    \begin{tabular}{llcccccc}
        \toprule
        \textbf{Ratings} & \textbf{Book Names}  & \textbf{Median Jaccard} & \textbf{Median Cosine} & \textbf{Median Levenshtein} & \textbf{Median Sequence Matcher} & \textbf{Median BLEU} & \textbf{Median ROUGE-L}\\
        \midrule
        2,735,023 & Romeo and Juliet & \textbf{0.762} & 0.906 & 0.931 & 0.957 & 0.708 & 0.941 \\ 
        831,152 & Siddhartha & 0.233 & 0.342 & 0.413 & 0.494 & 0.175 & 0.375 \\
        546,286 & The Time Machine & 0.250 & 0.345 & 0.432 & 0.506 & 0.228 & 0.392 \\
        492,129 & The Wonderful Wizard of Oz & 0.333 & 0.456 & 0.496 & 0.593 & 0.316 & 0.500 \\
        413,400 & Alice's Adventures in Wonderland & \textbf{0.913} & 0.962 & 0.944 & 0.963 & 0.931 & 0.978 \\
        362,694 & Peter Pan & 0.182 & 0.257 & 0.356 & 0.427 & 0.108 & 0.292 \\
        290,524 & Candide & 0.267 & 0.362 & 0.433 & 0.536 & 0.192 & 0.419 \\
        219,332 & The Tempest & 0.370 & 0.467 & 0.535 & 0.638 & 0.348 & 0.587 \\
        198,834 & Notes from the Underground & 0.235 & 0.301 & 0.402 & 0.478 & 0.168 & 0.364 \\
        183,818 & The Communist Manifesto & \textbf{0.952} & 0.980 & 0.961 & 0.971 & 0.953 & 0.980 \\
        169,009 & The Sign of the Four & 0.237 & 0.311 & 0.420 & 0.496 & 0.191 & 0.367 \\
        147,072 & The Legend of Sleepy Hollow & 0.407 & 0.526 & 0.597 & 0.667 & 0.435 & 0.585 \\
        140,217 & Through the Looking-Glass & 0.458 & 0.566 & 0.651 & 0.724 & 0.485 & 0.652 \\
        129,512 & The Island of Doctor Moreau & 0.158 & 0.236 & 0.321 & 0.387 & 0.046 & 0.263 \\
        48,922 & Just So Stories & 0.188 & 0.292 & 0.376 & 0.453 & 0.113 & 0.333 \\
        19,734 & Ethics & 0.343 & 0.456 & 0.488 & 0.591 & 0.279 & 0.500 \\
        7,582 & The Aesop for Children & 0.179 & 0.288 & 0.336 & 0.424 & 0.053 & 0.298 \\
        6,071 & The Secret of the Caves* & 0.103 & 0.186 & 0.263 & 0.333 & 0.014 & 0.196 \\
        787 & Simple Sabotage Field Manual & 0.114 & 0.156 & 0.268 & 0.334 & 0.018 & 0.195 \\
        344 & The First Book of Adam and Eve & 0.229 & 0.343 & 0.383 & 0.468 & 0.120 & 0.346 \\
        138 & The Philippines a Century Hence & 0.111 & 0.217 & 0.262 & 0.313 & 0.012 & 0.186 \\
        22 & The Emma Gees & 0.108 & 0.187 & 0.262 & 0.314 & 0.013 & 0.182 \\
        16 & Bab Ballads and Savoy Songs & 0.100 & 0.167 & 0.304 & 0.387 & 0.015 & 0.200 \\
        5 & Dragon Moon* & 0.088 & 0.160 & 0.244 & 0.309 & 0.012 & 0.170 \\
        4 & The Hallowell Partnership & 0.079 & 0.130 & 0.244 & 0.304 & 0.011 & 0.160 \\
        2 & Judas Ram & 0.088 & 0.145 & 0.246 & 0.304 & 0.012 & 0.163 \\
        2 & Rosin the Beau & 0.108 & 0.157 & 0.261 & 0.317 & 0.012 & 0.174 \\
        0 & Christina and the Boys* & 0.093 & 0.151 & 0.252 & 0.321 & 0.012 & 0.174 \\
        0 & Pegasus* & 0.111 & 0.204 & 0.261 & 0.329 & 0.014 & 0.196 \\
        0 & A girl and her ways* & 0.098 & 0.141 & 0.252 & 0.310 & 0.012 & 0.174 \\
        0 & Upside Down or Backwards & 0.073 & 0.137 & 0.242 & 0.306 & 0.011 & 0.158 \\
        0 & Science and Medieval Thought & 0.100 & 0.216 & 0.257 & 0.309 & 0.013 & 0.186 \\

        \bottomrule
    \end{tabular}%
    }
    \caption{Median similarity scores for books extracted from
Llama 3.1-instruct SFT-1000 samples. * denotes a book released \textbf{after} the knowledge cutoff date (December 2023).}
    \label{tab:book_testing_scores}
\end{table*}

\section{Analysis of Weight Updates}
\label{sec:weight-updates}


\begin{figure}[H]
    \centering
\resizebox{0.8\columnwidth}{!}{
\begin{tikzpicture}[
  layer/.style={rectangle, draw, minimum width=3.2cm, minimum height=1cm, rounded corners=2pt, align=center},
  lora/.style={rectangle, draw=red!70!black, thick, minimum width=3.2cm, minimum height=0.8cm, rounded corners=2pt, align=center, fill=red!10},
  group/.style={draw=black!50, dashed, thick, rounded corners=4pt, inner sep=5pt},
  arrow/.style={-{Latex}, thick},
  >=Latex
]

\node[layer] (ln1) {LayerNorm};

\node[layer] (attn) [below=1.5cm of ln1] {Self-Attention};

\node[lora] (qproj) [right=1cm of attn, yshift=1.38cm] {\small q\_proj (LoRA)};
\node[lora] (kproj) [below=0.1cm of qproj] {\small k\_proj (LoRA)};
\node[lora] (vproj) [below=0.1cm of kproj] {\small v\_proj (LoRA)};
\node[lora] (oproj) [below=0.1cm of vproj] {\small o\_proj (LoRA)};

\node[layer] (add1) [below=1.5cm of attn] {Residual + LayerNorm};

\node[layer] (mlp) [below=1.2cm of add1] {Feedforward (MLP)};

\node[lora] (gateproj) [right=1cm of mlp, yshift=0.92cm] {\small gate\_proj (LoRA)};
\node[lora] (upproj) [below=0.1cm of gateproj] {\small up\_proj (LoRA)};
\node[lora] (downproj) [below=0.1cm of upproj] {\small down\_proj (LoRA)};

\node[layer] (add2) [below=1.2cm of mlp] {Residual};

\draw[arrow] (ln1) -- (attn);
\draw[arrow] (attn) -- (add1);
\draw[arrow] (add1) -- (mlp);
\draw[arrow] (mlp) -- (add2);

\node[group, fit=(qproj) (kproj) (vproj) (oproj), label=above:{\small Self-Attention LoRA adapters}] (attnGroup) {};
\node[group, fit=(gateproj) (upproj) (downproj), label=above:{\small MLP LoRA adapters}] (mlpGroup) {};

\draw[dashed] (attn.east) -- ($(attnGroup.west)$);
\draw[dashed] (mlp.east) -- ($(mlpGroup.west)$);

\end{tikzpicture}
}
    \caption{Llama Transformer block with the LoRA adapters we train here. Only the specific layers inside Self-Attention and MLP blocks receive LoRA updates.}
    \label{fig:llama-lora-block}
\end{figure}
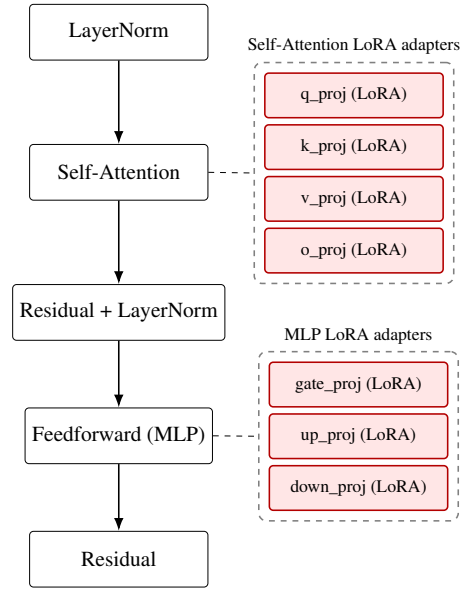

\end{document}